\documentclass{article}
\usepackage{spconf,amsmath,amsfonts,amssymb,graphicx,booktabs,multirow,url,flushend}

\title{PathAnchor: Path-Structured Evidence for Scientific Agents}

\name{\shortstack{Qiuhui Chen$^{1}$, Jiafan Lu$^{1}$, Shuaimin Tang$^{2}$, Tao Dai$^{1}$,\\
Suyuan Wang$^{1}$, Chenrui Ji$^{1}$, Zhenglei Zhou$^{3}$, and Weimin Zhong$^{1}$\sthanks{Corresponding author: Weimin Zhong.}}}
\address{$^{1}$School of Information Science and Engineering, East China University of Science and Technology\\
$^{2}$College of Intelligent Robotics and Advanced Manufacturing, Fudan University \quad $^{3}$Tencent}

\begin{document}
\ninept
\maketitle

\begin{abstract}
Scientific agents can retrieve relevant passages yet still lose functional order, mix evidence across sources, or state conclusions that exceed the retrieved record. We introduce PathAnchor, a bounded scientific reasoning system built on path-structured evidence workspaces. Instead of treating passages or extracted concepts as independent units, the system retrieves source-linked Material--Sensor--Signal--System trajectories that preserve role, direction, and the evidence supporting each transition. A controller uses three read-only tools to search paper-specific trajectories, trace paths across candidate sources, and open exact evidence before producing a claim-cited answer and an explicit evidence boundary. On 120 single- and cross-paper flexible-sensor questions, PathAnchor scores 82.6\% and leads six evaluated systems. Under a matched controller, corpus, and six-call budget, replacing unordered concept graphs with path-structured records raises source recall from 61.3\% to 82.9\%, increases answers whose claims all cite opened evidence from 69.2\% to 90.0\%, and reduces tool calls. These results show that evidence organization affects retrieval and citation completeness under fixed agent resources.
\end{abstract}


\section{Introduction}
\label{sec:intro}

Retrieval-augmented generation gives language models access to external knowledge, but access does not guarantee faithful scientific synthesis~\cite{lewis2020rag,izacard2023atlas}. A scientific answer often depends on a sequence of linked facts: a material realizes a sensing element, the element produces a measurable signal, and that signal enables a system-level function. When these facts are retrieved as isolated passages, an agent must reconstruct their roles and order during generation. It may combine incompatible sources, overlook a missing transition, or present a plausible completion as reported evidence.

Scientific question-answering systems combine evidence selection, iterative search, and citation-aware synthesis. PaperQA, LitQA2, and SciRAG use agentic document retrieval~\cite{lala2024paperqa,skarlinski2024language,ding2026scirag}; OpenScholar adds self-feedback and citation-aware synthesis~\cite{asai2026synthesizing}; and AstaBench and SciCUEval broaden evaluation~\cite{bragg2026astabench,yu2026scicueval}. Paper-grounded QA and claim-verification benchmarks measure answer--evidence alignment~\cite{dasigi2021qasper,wadden2020scifact,wadden2022scifactopen}. Structured extraction recovers entities and relations~\cite{luan2018scierc,jain2020scirex,dagdelen2024structured,caufield2024spires,li2026reconstructing}, while scientific knowledge graphs connect materials, properties, and applications~\cite{venugopal2024matkg,ke2024strainkg}. Yet a practical question remains: under matched evidence access, does organization change reasoning faithfulness and efficiency?

\begin{figure}[!t]
    \centering
    \includegraphics[width=0.98\columnwidth]{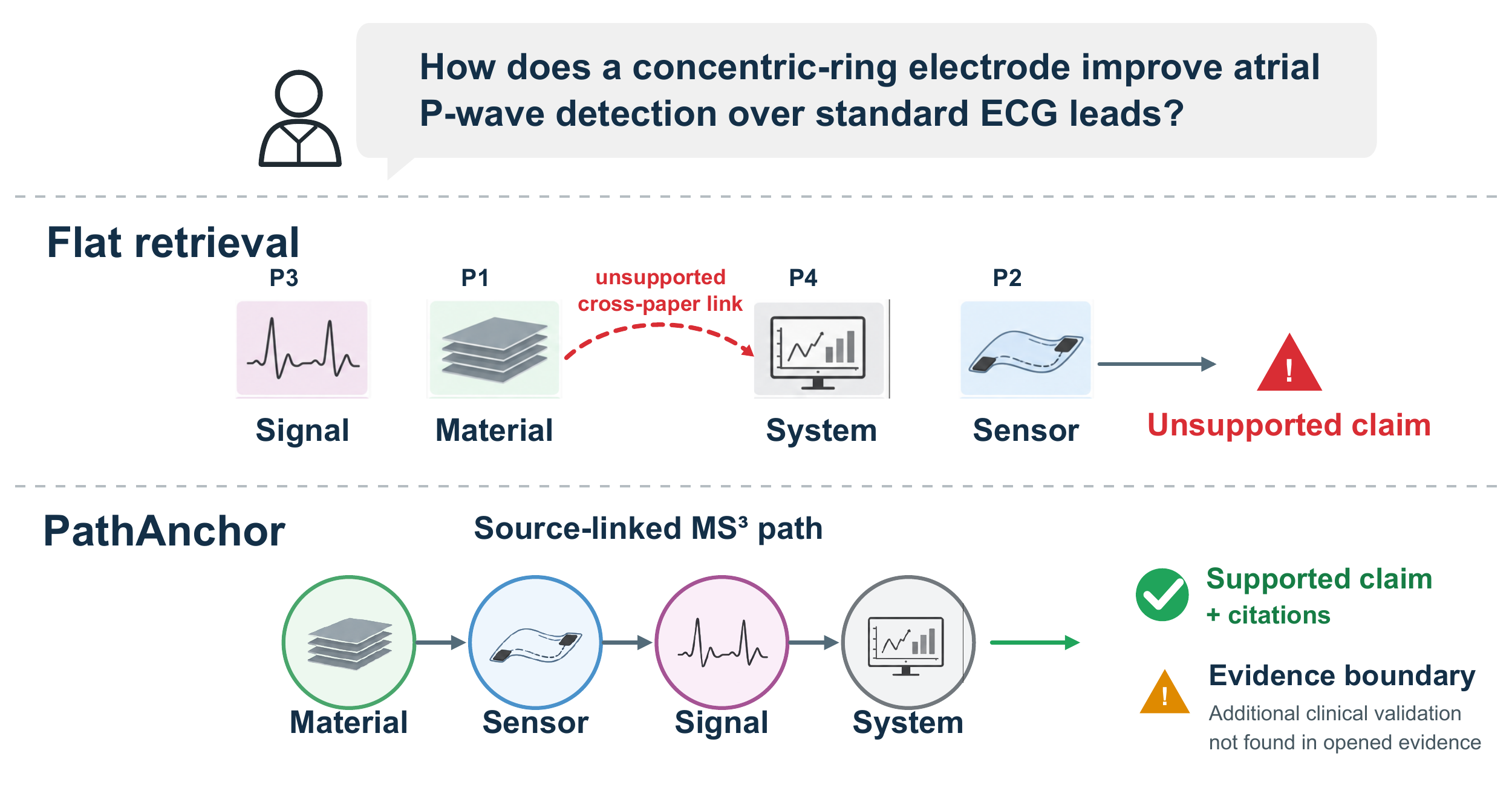}
    \caption{Illustrative comparison. P1--P4 denote distinct source papers. Flat retrieval can create an unsupported cross-paper link; PathAnchor preserves source-linked functional order, returns a cited supported claim, and explicitly marks evidence not found in the opened record.}
    \label{fig:motivation}
\end{figure}

We study this question using Material--Sensor--Signal--System (MS$^3$) functional trajectories. Each record is ordered, explicitly typed, and linked to source evidence. PathAnchor conditions retrieval on a scientific question, collects selected trajectories and opened passages in an inspectable workspace, and separates supported conclusions from an explicit evidence boundary. The agent uses three read-only tools and at most six calls.

We isolate evidence organization with a matched ablation that fixes the controller, corpus, evidence access, prompts, and call budget. On 120 FS EvidenceQA questions, path records improve source recall and citation-provenance completeness by 21.6 and 20.8 points while reducing tool use.

\begin{figure*}[!t]
    \centering
    \includegraphics[width=0.98\textwidth,trim=0 70 0 80,clip]{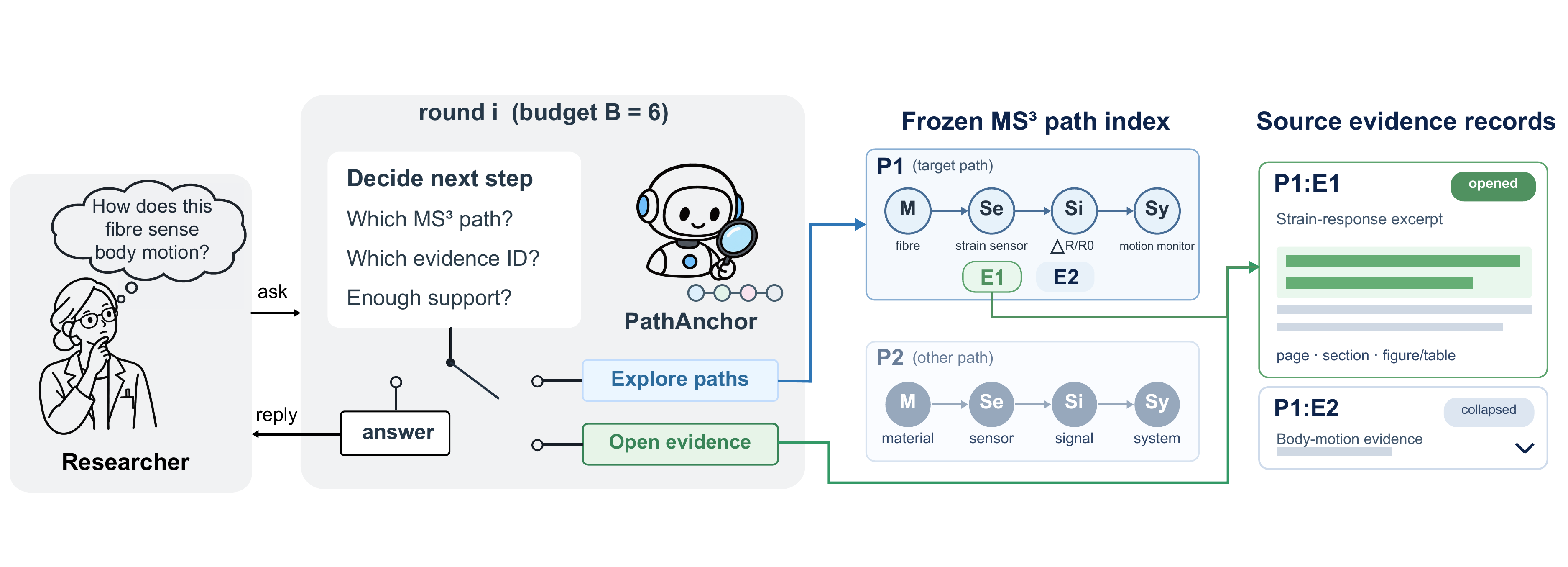}
    \caption{PathAnchor architecture for FS EvidenceQA. A single controller navigates paper-specific MS$^3$ paths, opens linked article records, and returns claim-cited output with an explicit evidence boundary. E1/E2 are schematic; only opened article records count as final citations.}
    \label{fig:overview}
\end{figure*}

\begin{table*}[!t]
\centering
\caption{Abridged PathAnchor prompt and two logged cases. The single-paper case uses \texttt{P\_081737}~\cite{xu2021coresheath}; the cross-paper case uses \texttt{P\_095797} and \texttt{P\_026428}~\cite{nasrollahzadeh2024mouthguard,sun2019zno}. Each \texttt{paper\_id:E\#} denotes an opened article record.}
\label{tab:prompt_example}
\small
\renewcommand{\arraystretch}{1.08}
\begin{tabular}{@{}p{\textwidth}@{}}
\toprule
\textbf{System prompt --- PathAnchor} \\
\midrule
\textbf{Tools and budget.} Begin without candidate papers. Use \texttt{search\_ms3}, \texttt{trace\_paths}, and \texttt{inspect\_evidence} with at most six calls.\par
\textbf{Evidence rule.} MS$^3$ path and mechanism IDs route retrieval but are not final proof. Every paper-specific claim cites an opened \texttt{paper\_id:E\#} article record; mark requests lacking support as insufficient. \\
\midrule
\textbf{Single-paper case --- \texttt{nv3\_single\_01\_01}} \\
\midrule
\textbf{Question.} For a SWCNT--RGO/PU fibre, how do TPU sheathing and 50 wt\% RGO affect sensing, and what was not validated?\par
\textbf{Actions.} Search $\rightarrow$ trace $(P\_081737)$ $\rightarrow$ inspect $(E1\text{--}E6)$.\par
\textbf{Answer.} After 60 min of ultrasonication, $\Delta R/R_0$ rose by 10 with TPU versus 690 without [E4]. For 50 wt\% RGO, gauge factors were 8.76, 43.8, and 114.7 over 1--150\%, 150--300\%, and 300--465\% strain [E3], with more than 1000 cycles [E6]. Opened records do not establish performance beyond 465\% strain or for other RGO compositions. \texttt{PARTIALLY\_SUPPORTED}. \\
\midrule
\textbf{Cross-paper case --- \texttt{nv3\_cross\_01\_08}} \\
\midrule
\textbf{Question.} Can a protective mouthguard also report deformation using a ZnO/GNP strain film?\par
\textbf{Actions.} Search twice $\rightarrow$ trace four candidate papers $\rightarrow$ inspect eight article records.\par
\textbf{Answer.} A mouthguard reduced transmitted incisor strain by over 55\% [\texttt{P\_095797:E5}]; a separate ZnO/GNP film measured strain over a 44\% range [\texttt{P\_026428:E5}]. Their paper-specific paths keep these results distinct: integration is a proposed design, not a tested device. \texttt{PARTIALLY\_SUPPORTED}. \\
\bottomrule
\end{tabular}
\end{table*}

\section{Path-Structured Scientific Agent}
\label{sec:method}

\subsection{Evidence-linked functional trajectories}

Let $\mathcal{D}$ be a frozen scientific corpus. Conventional passage retrieval represents a result as text and provenance. An MS$^3$ record additionally represents its functional organization as
\begin{equation}
p=(\mathcal V_p,\mathcal R_p,\mathcal A_p),
\end{equation}
where $\mathcal V_p$ contains typed Material, Sensor, Signal, and System nodes, $\mathcal R_p$ contains supported directed transitions, and $\mathcal A_p$ maps nodes and transitions to source evidence identifiers. The PathAnchor experiments retrieve complete four-role paths.

The domain evaluation uses a frozen corpus of 13,689 full-text conductive-fibre flexible-sensor papers, containing 26,648 trajectory records linked to 131,083 evidence items. A source-grounded construction pipeline first extracts compact evidence records, then assigns paper-specific MS$^3$ terms, and finally assembles only schema-valid relations supported by accepted evidence. Each evidence item retains its paper identifier, article excerpt, page, section, and available figure or table location. The resulting corpus and precomputed trajectory and passage indexes form the \emph{frozen MS$^3$ evidence base} in Fig.~\ref{fig:overview}. It is built once before evaluation; during a run, the controller can read but cannot modify it.

Trajectories and evidence passages are indexed separately with SQLite FTS5. Given query $q$, candidate-paper retrieval combines BM25 scores~\cite{robertson2009bm25} from the two views,
\begin{equation}
s(d\mid q)=\tfrac{1}{2}s_{\rm path}(d\mid q)+\tfrac{1}{2}s_{\rm ev}(d\mid q),
\end{equation}
and ranks paths within a paper by overlap between normalized query terms and the trajectory statement and fields. Each returned path exposes its roles, mechanism statement, and linked evidence identifiers. Thus path retrieval proposes a functional explanation, while opened article evidence remains the authority for final claims.

\subsection{Bounded evidence workspace}

The agent begins each question with an empty workspace $W_0$ and may issue three read-only actions:
\begin{itemize}
    \item \texttt{search\_ms3}: retrieve candidate trajectories and evidence identifiers;
    \item \texttt{trace\_paths}: return stored complete paths and short evidence previews from papers already visible in the workspace;
    \item \texttt{inspect\_evidence}: open exact source records, including article excerpt, page, section, and figure or table location.
\end{itemize}
The search action returns at most eight candidate trajectories, path tracing returns at most six paths, and evidence inspection opens at most 12 identifiers per call. After action $a_t$, the workspace is updated as $W_t=W_{t-1}\cup O(a_t)$, where $O(a_t)$ is the returned observation. Search terminates when the workspace supports the requested answer or reaches budget $B=6$. The controller never receives the reference answer, hidden label, or evaluation rubric.

The draft answer links each substantive paper-specific claim $c_j$ to opened evidence $\mathcal A_j$ and returns a separate boundary statement $b$,
\begin{equation}
y=\{(c_j,\mathcal A_j)\}_{j=1}^{J}\cup\{b\},\qquad
\mathcal A_j\subseteq E(W_T).
\end{equation}
The controller prompt asks it to distinguish demonstrated evidence from proposals and to mark unsupported requests in $b$. When findings from different papers enter one workspace, each path and evidence record retains its paper identifier. The controller can align their functional roles in an answer while citing each measured property to its own source; a proposed cross-paper connection belongs in the boundary statement, not in either stored path. A deterministic provenance audit checks each citation against opened records, while the answer evaluation assesses grounding. This claim-level design follows work on citation support, verified quotations, and atomic factual precision~\cite{gao2023citations,menick2022quotes,zhang2025longcite,min2023factscore}. The prompt limits outputs to 300 words, seven claims, ten unique citations, and ten quantitative facts. Tool calls, workspace states, and final answers are retained for audit.

\begin{table}[t]
\centering
\caption{Overall scores (\%); FS Unsup. is the judge-rated count per answer. Ours denotes task-adapted MS$^3$ workflows on the three public tasks and PathAnchor on FS EvidenceQA. DSeek, PQA2, and OSch. are DeepSeek V4 Pro + Search, PaperQA2, and OpenScholar.}
\label{tab:main}
\small
\setlength{\tabcolsep}{2.0pt}
\renewcommand{\arraystretch}{1.02}
\begin{tabular}{lcccccc}
\toprule
Metric & DSeek & Kimi & GLM & PQA2 & OSch. & \textbf{Ours} \\
\midrule
FS All & 66.4 & 77.2 & 72.0 & 67.9 & 55.2 & \textbf{82.6} \\
FS Single & 78.8 & 87.6 & 84.4 & 76.0 & 60.0 & \textbf{91.9} \\
FS Cross & 54.1 & 66.9 & 59.7 & 59.8 & 50.4 & \textbf{73.3} \\
FS Unsup.$\downarrow$ & 1.917 & 1.525 & 1.683 & 2.217 & 2.908 & \textbf{1.267} \\
\midrule
LQA & 32.0 & 33.3 & 16.0 & 4.0 & 17.3 & \textbf{44.0} \\
SQA & 31.9 & 49.5 & 43.5 & 31.1 & 54.0 & \textbf{66.0} \\
SciCU & 91.0 & 84.0 & 91.5 & 40.5 & 48.5 & \textbf{98.0} \\
\bottomrule
\end{tabular}
\end{table}

These access rules make each final citation traceable to an article record opened during the run.

\subsection{Matched representation control}

To isolate information organization, we construct Role Graph-Agent with the same controller, source corpus, prompts, evidence passages, and six-call budget. Its tools return extracted concepts, paper provenance, evidence identifiers, and source excerpts, but withhold the MS$^3$ role labels, directed relation types, path order, and mechanism statements. The experiment therefore changes the structure presented to the agent without changing the underlying evidence it can access.

\section{Experimental Protocol}
\label{sec:experiments}

\subsection{Benchmarks}

We evaluate complementary public and domain-controlled tasks. \textbf{LitQA2-FullText} contains all 75 test questions and is scored by exact option-label accuracy~\cite{skarlinski2024language}. \textbf{ScholarQA-CS2} contains all 100 cases and uses the official global average of answer coverage, relevance, citation recall, and citation precision~\cite{bragg2026astabench}. \textbf{SciCUEval Materials} contains 200 questions, with 25 from each combination of two input modalities and four scientific competencies; we report the eight-stratum macro-average~\cite{yu2026scicueval}.

Our fourth benchmark, \textbf{FS EvidenceQA}, comprises 120 frozen, evidence-derived questions grounded in conductive-fibre flexible-sensor papers: 60 single-paper and 60 cross-paper questions balanced over six categories. A generator saw anonymized article excerpts but no paper identifiers, MS$^3$ role fields, retrieval outputs, or method ranks, and produced questions, answer outlines, and unsupported-content boundaries. Answers receive 0--4 ratings for correctness, grounding, evidence completeness, and boundary control; their mean is normalized to percentage. Method identities are blinded to the Qwen3-30B-A3B evaluator.

The source sampling excluded papers from an earlier 20-question retrieval audit, and source papers are disjoint across questions. The model-generated questions make FS EvidenceQA a controlled diagnostic of source-grounded answering.

\subsection{Systems and evaluation}

We compare six configurations under controlled source access: direct Evidence/Search with DeepSeek V4 Pro, Kimi K3, or GLM-5.2; PaperQA2~\cite{lala2024paperqa}; OpenScholar with external retrieval disabled~\cite{asai2026synthesizing}; and our DeepSeek V4 Pro-based systems. FS EvidenceQA and the matched representation test use PathAnchor with the frozen four-role path index in Sec.~\ref{sec:method}. The three public tasks use task-adapted MS$^3$ workflows and native answer formats: LitQA2 and ScholarQA draw on their benchmark evidence packets, while SciCUEval supplies item-level text and tables. Failed or truncated outputs are scored incorrect rather than discarded. FS EvidenceQA uses paired 10,000-replicate question-level bootstrap intervals. The representation ablation additionally measures hidden-gold source recall in the final workspace, answers whose claims all cite opened evidence, claims without opened-evidence citations, and tool calls.

The public evaluations use their native endpoints rather than a shared judge. LitQA2 is exact option-label accuracy; ScholarQA uses the official section, inline-citation, and supporting-snippet format and averages coverage, relevance, citation recall, and citation precision; SciCUEval uses exact-answer or absence-rejection accuracy and a macro-average over its eight modality--competency strata. Bootstrap contrasts resample the six answers to each FS EvidenceQA question together, preserving paired comparisons; SciCUEval resampling preserves its eight strata.

A separate retrieval diagnostic compares full-text evidence FTS against MS$^3$ mechanism-field FTS on the same frozen corpus. It uses all 180 frozen questions (including the 120 FS EvidenceQA cases). Candidate relevance is graded 0--3 by a DeepSeek V4 Pro proxy judge blinded to retrieval method and rank; nDCG@10 uses paired question-level bootstrap intervals.

\begin{figure*}[t]
    \centering
    \begin{minipage}[t]{0.555\textwidth}
        \vspace{0pt}
        \centering
        \includegraphics[width=\linewidth]{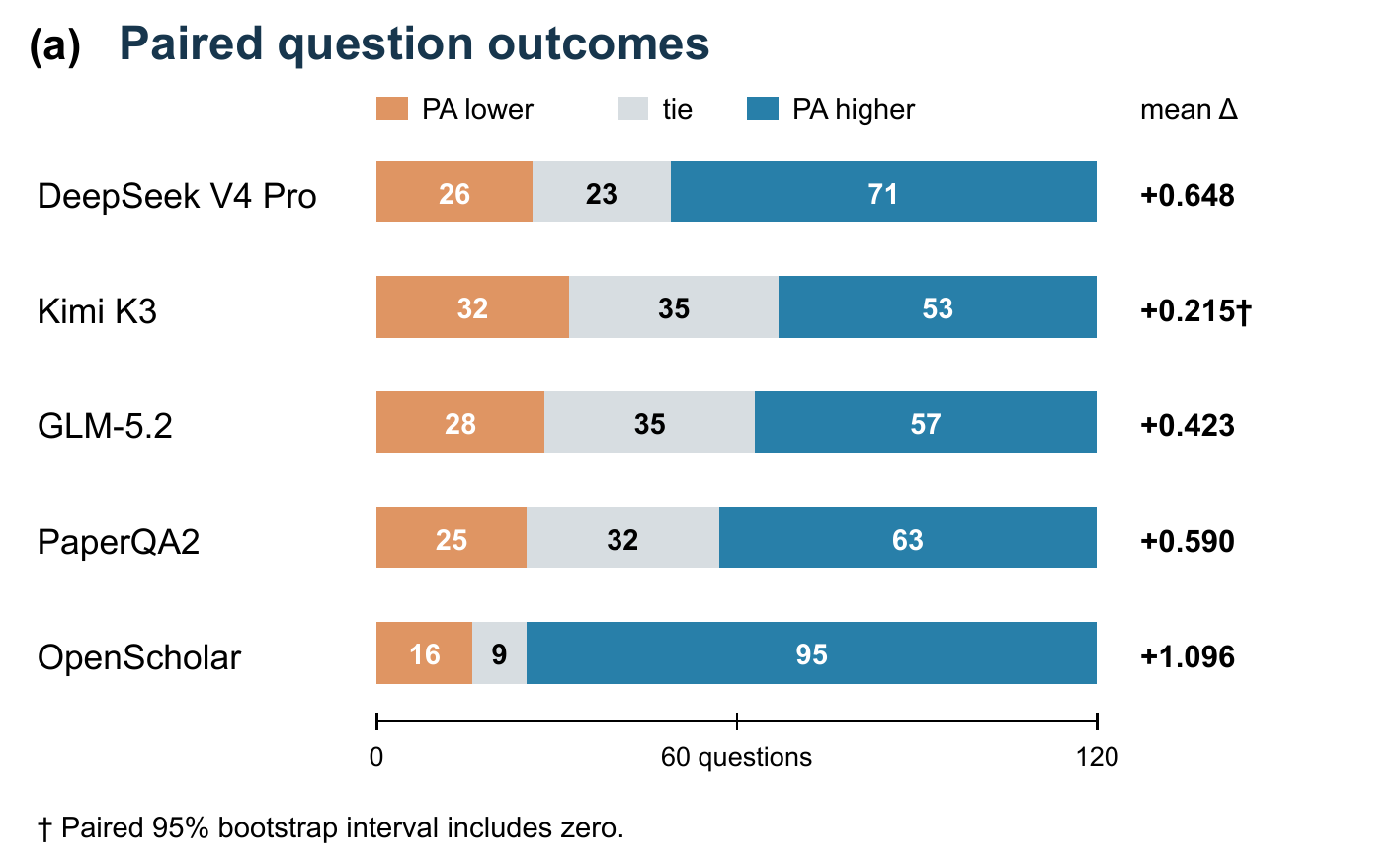}
    \end{minipage}\hfill
    \begin{minipage}[t]{0.415\textwidth}
        \vspace{0pt}
        \centering
        \includegraphics[width=\linewidth]{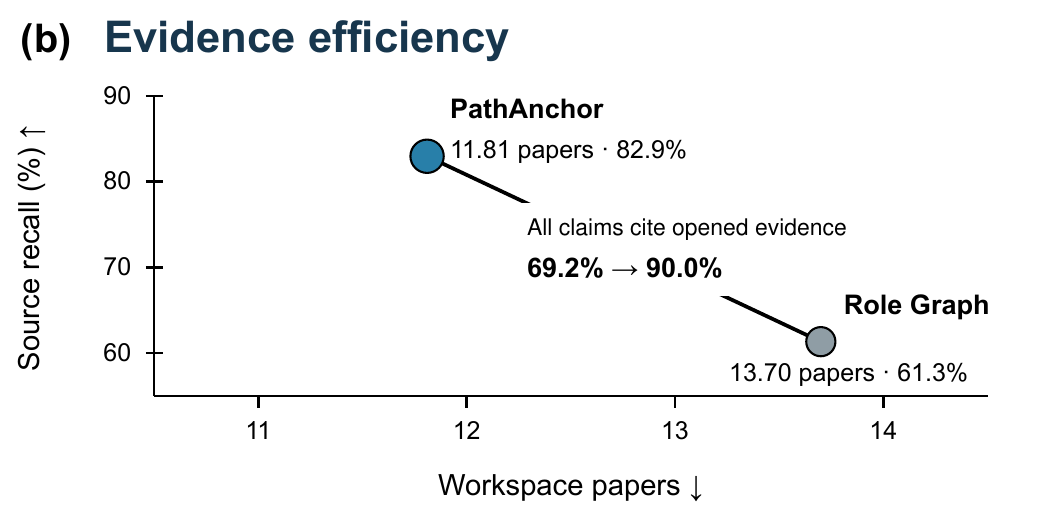}\\[-0.25em]
        \includegraphics[width=\linewidth]{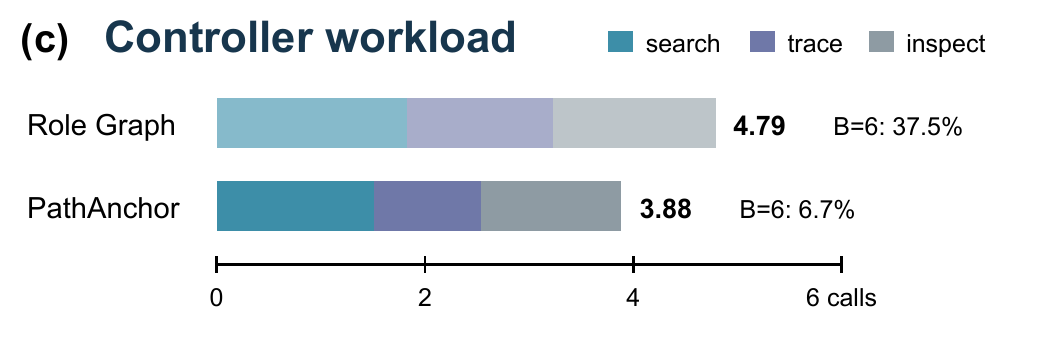}
    \end{minipage}
    \caption{Paired outcomes and matched representation control. (a) FS EvidenceQA: PathAnchor (PA) versus each baseline; $\dagger$ marks a paired 95\% interval crossing zero. (b) Source recall versus workspace size, with the proportion of answers whose claims all cite opened evidence. (c) Mean calls by action; $B{=}6$ is the budget-hit rate.}
    \label{fig:results-analysis}
\end{figure*}

\section{Results and Analysis}
\label{sec:results}

\subsection{Multi-benchmark performance}

On FS EvidenceQA, PathAnchor reaches 82.6\% (3.304/4) and leads both question types. Its 0.215-point margin over Kimi K3 has a paired 95\% bootstrap interval of $[-0.033,0.469]$; advantages over the other four systems exclude zero.

Cross-paper scores fall for every system, but PathAnchor remains highest (2.933 versus 3.675 on single-paper cases; Kimi K3: 2.675 versus 3.504). Unsupported claims increase from 0.583 to 1.950 per PathAnchor answer (Kimi K3: 0.817 to 2.233), underscoring the difficulty of cross-source synthesis.

The task-adapted MS$^3$ workflows rank first on the three public evaluations (Table~\ref{tab:main}), exceeding the strongest comparator by 10.7, 12.0, and 6.5 points on LitQA2-FullText, ScholarQA-CS2, and SciCUEval Materials, respectively.

\subsection{Claim-level quality and uncertainty}

PathAnchor's correctness, grounding, evidence-completeness, and boundary-control scores are 3.325, 3.300, 3.308, and 3.283, respectively; it also produces the fewest unsupported claims (1.267 per question). Fig.~\ref{fig:results-analysis}a shows paired advantages over all comparators, with Kimi K3 the only interval crossing zero.

The cross-paper case in Table~\ref{tab:prompt_example} illustrates the distinction between combining evidence and claiming an integrated device. One source measures the mouthguard's impact attenuation; another measures the ZnO/GNP film's strain response. PathAnchor cites those observations under separate paper IDs and describes the sensing mouthguard as a proposed combination. The answer therefore addresses the cross-paper question without transferring the film's measurement to the mouthguard.

\subsection{What path structure changes}

\begin{table}[t]
\centering
\caption{Retrieval nDCG@10 (method-blind model relevance). $\Delta$ is MS$^3$ minus text FTS with paired 95\% CI.}
\label{tab:retrieval}
\small
\setlength{\tabcolsep}{1.0pt}
\renewcommand{\arraystretch}{1.04}
\begin{tabular}{lrrrl}
\toprule
Question stratum & $n$ & Text & MS$^3$ & $\Delta$ [95\% CI] \\
\midrule
Covered, all & 120 & .688 & .754 & +.066 [.021, .112] \\
\quad Single-paper & 60 & .772 & .814 & +.042 [-.022, .108] \\
\quad Cross-paper & 60 & .604 & .693 & +.089 [.027, .152] \\
Partially covered & 30 & .765 & .740 & -.024 [-.093, .039] \\
Out-of-domain & 30 & .559 & .587 & +.028 [-.035, .092] \\
\bottomrule
\end{tabular}
\end{table}

The retrieval-only test in Table~\ref{tab:retrieval} provides a complementary view: MS$^3$ improves ranking on the 120 covered cases, especially cross-paper questions, while the partially covered and out-of-domain contrasts have intervals spanning zero. The cross-paper gain (+.089 nDCG@10) exceeds the single-paper gain (+.042), matching the setting in which the agent must locate distinct sources and align their functional contributions.

In the central matched test (Fig.~\ref{fig:results-analysis}b--c), path structure raises source recall by 21.6 points and the rate of answers whose claims all cite opened evidence from 69.2\% to 90.0\%, although both agents can open the same passages. Claims without opened-evidence citations fall from 0.450 to 0.133 per answer, and calls fall by 0.91. The ordered representation therefore concentrates answer-relevant evidence: the controller finds more gold sources with fewer actions and produces more complete citation provenance.

All 120 matched PathAnchor runs completed successfully. PathAnchor exposed fewer papers than Role Graph (11.81 versus 13.70) and achieved higher citation validity (99.90\% versus 96.39\%). It reached the six-call limit in 6.7\% of runs versus 37.5\% for Role Graph, consistent with earlier stopping once source-specific support was found.

Together, higher recall with fewer exposed papers and tool calls shows that ordered paths focus the controller on answer-relevant evidence. The representation ablation changes role labels, relation direction, path order, and mechanism statements as a bundle.

\subsection{Discussion}

The retrieval and agent tests locate the benefit at two stages. The path index improves ranking on covered questions, and the matched controller retrieves more gold sources despite a smaller workspace. In Table~\ref{tab:prompt_example}, those sources remain separately attributable when the question asks about a combined design. Source-local functional order thus supports both evidence selection and cross-paper synthesis under a limited call budget. Future work will test partial paths and individual representation components across additional controllers and expert-authored questions.

\section{Conclusion}

PathAnchor organizes scientific evidence as ordered, source-linked MS$^3$ trajectories under a six-call budget. On FS EvidenceQA it achieves the highest observed score among six tested systems. The matched ablation shows that path structure improves source recall and opened-evidence citation provenance while reducing tool calls. An explicit evidence boundary makes unsupported requests visible, while final citations remain tied to opened article records.

\clearpage

\noindent\textbf{Acknowledgment.} This work was supported in part by the Fundamental and Interdisciplinary Disciplines Breakthrough Plan of the Ministry of Education of China (JYB2025XDXM402) and the National Natural Science Foundation of China (Nos. 62603270 and 62633009).

\vspace{0.2em}
\noindent\textbf{Compliance with Ethical Standards.} This study involved neither human participants nor animals; ethical approval was not required. The authors declare no relevant financial or nonfinancial conflicts of interest.

\bibliographystyle{IEEEbib}
\bibliography{refs}

\end{document}